\documentclass{article}

\usepackage{PRIMEarxiv}

\usepackage[utf8]{inputenc} 
\usepackage[T1]{fontenc}    
\usepackage{hyperref}       
\usepackage{url}            
\usepackage{booktabs}       
\usepackage{amsfonts}       
\usepackage{nicefrac}       
\usepackage{microtype}      
\usepackage{lipsum}
\usepackage{fancyhdr}       
\usepackage{graphicx}       
\graphicspath{{media/}}     
\usepackage{algorithm}
\usepackage{algpseudocode}
\usepackage{amsmath}
\usepackage{multirow}
\usepackage[normalem]{ulem}
\useunder{\uline}{\ul}{}

\title{PathVLM-R1: A Reinforcement Learning-Driven Reasoning Model for Pathology Visual-Language Tasks
}

\author{
  JIANYU WU\footnotemark[1] \\
  Fudan University \\
  China, Shanghai\\
  \texttt{jywu24@m.fudan.edu.cn} \\
   \And
  HAO YANG\footnotemark[1] \, \footnotemark[2] \\
  Fudan University \\
  China, Shanghai \\
  \texttt{21210720040@m.fudan.edu.cn} \\
   \And
  XINHUA ZENG\footnotemark[2] \\
  Fudan University \\
  China, Shanghai \\
  \texttt{zengxh@fudan.edu.cn} \\
  \AND
  GUIBING HE \\
  Fudan University \\
  China, Shanghai \\
  \texttt{gbhe23@m.fudan.edu.cn} \\
  \And
  ZHIYU CHEN \\
  Fudan University \\
  China, Shanghai \\
  \texttt{zhiyuchen23@m.fudan.edu.cn} \\
  \And
  ZIHUI LI \\
  Fudan University \\
  China, Shanghai \\
  \texttt{zihuili24@m.fudan.edu.cn} \\
  \And
  XIAOCHUAN ZHANG \\
  Fudan University \\
  China, Shanghai \\
  \texttt{xiaochuanzhang24@m.fudan.edu.cn} \\
  \And
  YANGYANG MA \\
  Fudan University \\
  China, Shanghai \\
  \texttt{12111240009@fudan.edu.cn} \\
  \And
  RUN FANG \\
  Fudan University \\
  China, Shanghai \\
  \texttt{fangrun@ywfudan.cn} \\
  \And
  YANG LIU \\
  Fudan University \\
  China, Shanghai \\
  \texttt{yang\_liu20@fudan.edu.cn} \\
}

\begin{document}
\maketitle
\renewcommand{\thefootnote}{\fnsymbol{footnote}}
\footnotetext[1]{These authors contributed equally to this work.}
\footnotetext[2]{Corresponding authors.}

\begin{abstract}
The diagnosis of pathological images is often limited by expert availability and regional disparities, highlighting the importance of automated diagnosis using Vision-Language Models (VLMs). Traditional multimodal models typically emphasize outcomes over the reasoning process, compromising the reliability of clinical decisions. To address the weak reasoning abilities and lack of supervised processes in pathological VLMs, we have innovatively proposed PathVLM-R1, a visual language model designed specifically for pathological images. We have based our model on Qwen2.5-VL-7B-Instruct and enhanced its performance for pathological tasks through meticulously designed post-training strategies. Firstly, we conduct supervised fine-tuning guided by pathological data to imbue the model with foundational pathological knowledge, forming a new pathological base model. Subsequently, we introduce Group Relative Policy Optimization (GRPO) and propose a dual reward-driven reinforcement learning optimization, ensuring strict constraint on logical supervision of the reasoning process and accuracy of results via cross-modal process reward and outcome accuracy reward. In the pathological image question-answering tasks, the testing results of PathVLM-R1 demonstrate a 14\% improvement in accuracy compared to baseline methods, and it demonstrated superior performance compared to the Qwen2.5-VL-32B version despite having a significantly smaller parameter size. Furthermore, in out-domain data evaluation involving four medical imaging modalities—Computed Tomography (CT), dermoscopy, fundus photography, and Optical Coherence Tomography (OCT) images—PathVLM-R1's transfer performance improved by an average of 17.3\% compared to traditional SFT methods. These results clearly indicate that PathVLM-R1 not only enhances accuracy but also possesses broad applicability and expansion potential.
\end{abstract}


\section{Introduction}
In medical pathology diagnosis, accurate analysis of pathological slides is crucial for clinical decision-making. However, the expertise of professional pathologists varies significantly, and the growing demand for diagnoses is not being adequately met. To alleviate disparities in diagnostic and treatment levels across regions and relieve the pressure on medical human resources, researchers have begun developing visual language models to assist in the diagnosis of pathological images.

In medical diagnostics, understanding the rationale behind the diagnosis is as important as the precision of the results. Currently, most visual language models depend on end-to-end training with supervised fine-tuning (SFT)\cite{PLip,pathchat,ALBEF}. These models aim to enhance task-specific performance by optimizing a loss function, often tending to memorize and mimic patterns in the data, which results in a lack of transparent and credible reasoning capabilities. Furthermore, SFT is heavily reliant on the quality and diversity of training data, which can lead to overfitting and poor performance on unseen data, a crucial concern in the medical field.

\begin{figure*}[h]
  \centering
  \label{fig:1}
  \includegraphics[width=\linewidth]{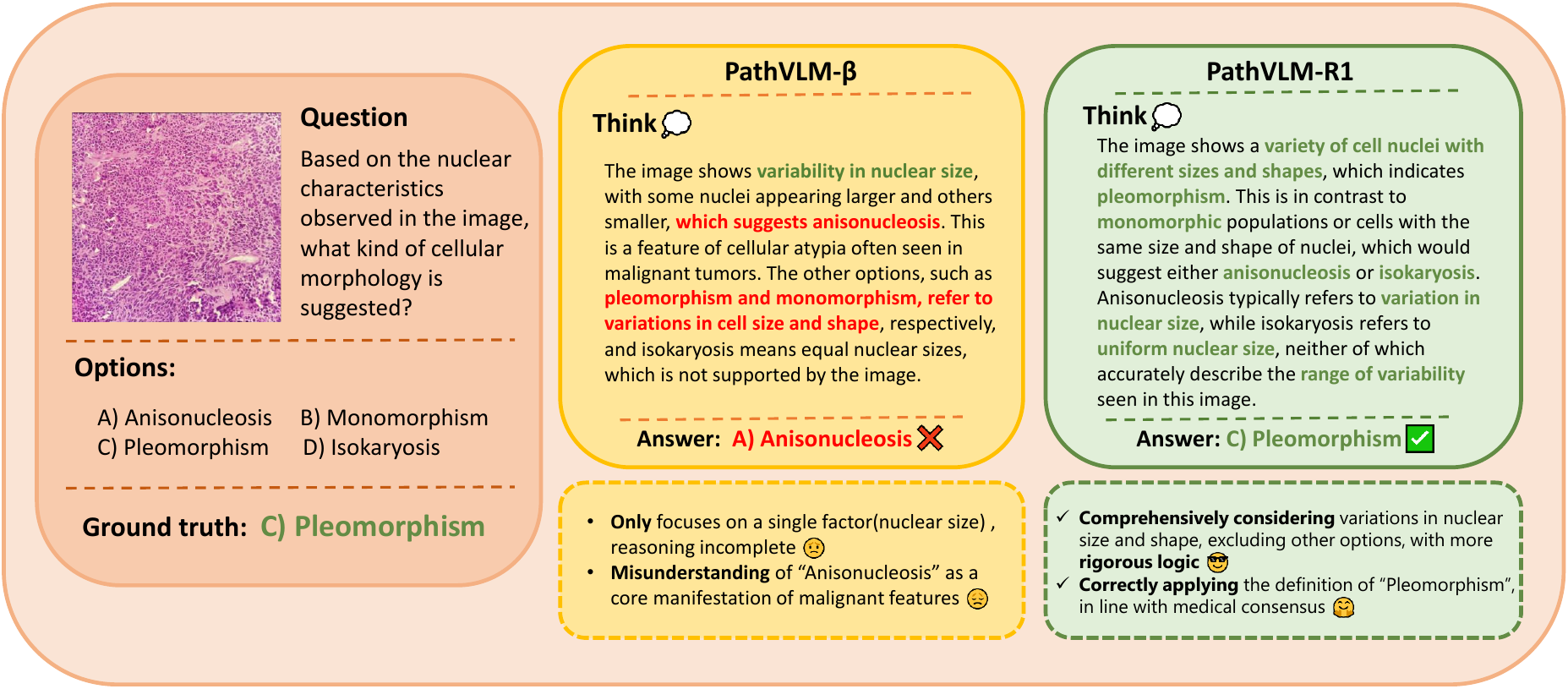}
  \caption{Comparison of the final PathVLM-R1 model and its variant PathVLM-$\beta$. For the same pathology image question-answering task, the PathVLM-R1, which incorporates cross-modal procedural rewards, demonstrates superior rigor in reasoning and accuracy in medical knowledge compared to the PathVLM-$\beta$, which utilizes only accuracy rewards and format rewards. Specifically, in this example, PathVLM-$\beta$ only focuses on differences in nuclear size while neglecting that "pleomorphism" includes multiple aspects of variation, such as nuclear shape and structure. In contrast, PathVLM integrates multiple nuclear features, accurately grasping the definition of "pleomorphism" with a more comprehensive thought process and stronger rigor.}
\end{figure*}

In contrast, reinforcement learning methods, widely applied in the training of reasoning models, offer a new approach to address these issues. By maximizing accumulated rewards, reinforcement learning encourages models to explore causal relationships between actions and outcomes, thereby constructing explainable reasoning steps and exhibiting stronger generalization abilities across various scenarios. However, directly applying reinforcement learning to foundational models in pathology tasks poses challenges:

Basic models lacking pathological knowledge may render initial reinforcement learning strategies random and ineffective. The model must learn both pathological knowledge and reasoning logic simultaneously, leading to sparse reward signals due to the absence of prior knowledge and requiring a large number of samples.

Traditional reinforcement learning methods, such as Proximal Policy Optimization (PPO)\cite{PPO}+Outcome Reward Model (ORM)\cite{ORM&PRM}, focus on the final outcome due to their black-box model structure, making it difficult to effectively verify the correctness of the reasoning process and thus falling short of the transparency and interpretability required for medical tasks. On the other hand, PPO\cite{PPO}+Process Reward Model (PRM)\cite{ORM&PRM} entails high annotation costs.

To address these challenges, we have designed a staged post-training process for the model: The first stage is the knowledge injection phase, where supervised fine-tuning on pathological data endows the model with basic pathological knowledge. The second phase constructs reasoning capabilities, where we innovatively introduce cross-modal process rewards to supervise the quality of the reasoning process from the perspectives of process completeness and pathological knowledge accuracy, in conjunction with results accuracy rewards to evaluate final answers, forming a "process-result" closed-loop supervision. Ultimately, the integration of cross-modal process rewards and results accuracy rewards is achieved through Group Relative Policy Optimization (GRPO)\cite{GRPO}, ensuring a balance between process rationality and outcome accuracy.

Our contributions include:
\begin{itemize}
\item A pathology image modality visual language model: We introduce PathVLM-R1, the first visual language model that supports pathology image modalities with reasoning ability, capable of providing detailed reasoning explanations and exhibiting excellent generalization performance.
\item An innovative dual reward mechanism: This transforms medical demands for process interpretability and result accuracy into learnable dual reward signals by quantifying reasoning step rationality through cross-modal process rewards and constraining diagnostic precision of the final answer through results accuracy rewards, ultimately achieving joint optimization of evidence derivation and conclusion generation.
\item Superior training and parameter efficiency: PathVLM-R1 demonstrates exceptional efficiency in terms of training parameters and sample quantity, achieving performance surpassing traditional supervised fine-tuning baselines on pathological images and other medical imaging modalities with only 7 billion parameters and 1000 samples for reinforcement learning optimization.
\end{itemize}
\section{Related Works}

\subsection{General and Medical Visual Language Models}

In recent years, general Visual Language Models such as LLama, Gemma, and QwenVL\cite{llama3,gemma2,qwen2.5,qwenvl,deepseekvl,llava} have achieved a deep alignment of visual and linguistic representations through pre-training on large-scale image and text data, significantly advancing the field of multimodal understanding technology. The underlying Transformer\cite{transformer} architecture and contrastive learning strategies provide effective means for capturing semantic associations between images and language. These technological advancements offer important theoretical and practical foundations for constructing medical visual language models.

In the medical domain, researchers have drawn on the successful experience of general VLMs by transferring these models to medical images and clinical texts. Through supervised fine-tuning and instruction tuning, they adapt the models for specialized medical tasks. Traditional medical image-assisted diagnosis methods\cite{Chexnet,lung,pubmedclip} require fine-tuning for each specific task. In contrast, the new learning paradigm using VLMs can effectively leverage web data and zero-shot prediction, allowing VLMs to identify new objects with few or no image samples\cite{VLMsur}. For example, the LLaVA-Med\cite{LLaVA-Med} models leverage large volumes of medical image-text pairs for further training, achieving significant progress in tasks like automated image report generation and visual question-answering. Despite these advancements, current medical visual language models still face several challenges:

Firstly, many medical visual language models primarily employ methods such as supervised fine-tuning, unsupervised pre-training, and contrastive learning. These methods aim to capture the statistical correlations and surface patterns between images and text. The commonly used training data typically involve paired sets of images and final diagnostic reports, clinical records, or radiology reports. Such data often only record diagnostic outcomes or report summaries, lacking detailed descriptions of the intermediate reasoning steps involved in the diagnostic process. This limitation affects existing medical visual language models in terms of inference capability and clinical interpretability, making it difficult to construct clear and transparent medical reasoning chains that could provide deep diagnostic insights.Recently, researchers have attempted to enhance model reasoning performance on medical tasks using reinforcement learning with GRPO as the optimization strategy. For instance, the MedVLM-R1\cite{medvlm-R1} model has surpassed baseline methods that use supervised fine-tuning in tests involving MRI, CT, and OCT modalities. However, it failed in tests on pathology modalities. Moreover, its approach of directly applying reinforcement learning to base models lacking fundamental knowledge has resulted in relatively limited performance improvements.

Secondly, since the training data is often restricted to specific domains and imaging modalities, the models have limited generalization capability when confronted with diverse clinical scenarios.

\subsection{Reinforcement Learning in the Post-training Phase}
Reinforcement Learning from Human Feedback (RLHF)\cite{RLHF} is a common practice during the post-training phase of large models\cite{gpt4,hybridflow,rlaif}. RLHF is a methodological approach that integrates human feedback into the reinforcement learning process. By collecting and analyzing human preferences or ratings for different behaviors, RLHF trains a reward model that simulates human preferences. Subsequently, using reinforcement learning algorithms such as Proximal Policy Optimization (PPO), this approach employs the initial model as a policy and uses the reward model's ratings as optimization signals to iteratively refine the model outputs, aligning them more closely with human values. A seminal work is InstructGPT\cite{instructgpt}, which employs Reinforcement Learning from Human Feedback (RLHF) to enhance the model's ability to adhere to instructions given in text prompts, thereby reducing harmful outputs and improving factual accuracy.

There are primarily two types of reward models: the Outcome Reward Model (ORM) and the Process Reward Model (PRM). The ORM focuses on the final output, rating and evaluating the overall result generated by the model, while the PRM scores each step in the generation process. Training these reward models typically relies on a substantial amount of human-labeled data. However, in fields such as medicine, where high-quality annotated data is scarce, the difficulty of training reward models increases significantly.

\section{Methods}
In the post-training phase for pathology tasks, we employ carefully crafted training strategies to enhance the model's reasoning capability and performance in these tasks. Below, we will introduce the specific optimization algorithms and training processes.
\subsection{Group Relative Policy Optimization (GRPO)}
In the medical field, the sensitivity of data, barriers due to specialized knowledge, and the high cost of annotations make acquiring high-quality reasoning process labels extremely difficult. This presents significant challenges to traditional reinforcement learning methods that rely on explicit reward models or value functions, such as Proximal Policy Optimization (PPO). In this context, GRPO emerges as a highly promising solution.

Group Relative Policy Optimization is an efficient reinforcement learning algorithm designed to enhance model reasoning by optimizing policy gradients through group-based relative advantage estimation and regularization, addressing limitations of traditional methods like PPO. The algorithm operates on a problem distribution $P(Q)$, where each problem $q$ is processed by an old policy $\pi_{\theta^{\text {old }}}$ to generate $G$ candidate responses $\left\{o_i\right\}_{i=1}^G$, with $\pi_\theta$ denoting the current policy to be optimized and $\pi_{\mathrm{ref}}$ serving as a frozen reference policy to constrain updates via KL divergence. For each candidate response $o_i$,  a rule-based reward $r_i$ is computed, and relative advantage is derived through group-wise normalization: 
\begin{align}
A_i=\frac{r_i-\operatorname{mean}\left(r_1, \ldots, r_G\right)}{\operatorname{std}\left(r_1, \ldots, r_G\right)}
\end{align} 
which measures the quality of $o_i$ relative to the group. The objective function of GRPO balances policy improvement and stability as:

\begin{align}
\mathcal{J}_{\text{GRPO}}(\theta) 
&= \mathbb{E}\Bigg[
q \sim P(Q),\ 
\left\{o_i\right\}_{i=1}^G \sim \pi_{\theta_{\mathrm{old}}}(O \mid q)
\Bigg] \nonumber \\
&\quad \frac{1}{G} \sum_{i=1}^G \Bigg\{
\min \Bigg[
\frac{\pi_\theta(o_i \mid q)}{\pi_{\theta_{\mathrm{old}}}(o_i \mid q)} A_i,\
\operatorname{clip} \Bigg(
\frac{\pi_\theta(o_i \mid q)}{\pi_{\theta_{\mathrm{old}}}(o_i \mid q)},\ 
1 - \varepsilon,\ 
1 + \varepsilon
\Bigg) A_i\Bigg] - \beta\ \mathbb{D}_{\mathrm{KL}}\left[\pi_\theta \| \pi_{\mathrm{ref}} \right]
\Bigg\}
\end{align}

where $G$ is the number of candidate responses per problem, $\frac{\pi_\theta\left(o_i \mid q\right)}{\pi_{\theta_{\text {old }}}\left(o_i \mid q\right)}$is the policy ratio reflecting the change in response generation probability, $\operatorname{clip}(\cdot)$ with hyperparameter $\epsilon$ stabilizes training by limiting extreme updates, and $\beta \mathbb{D}_{\mathrm{KL}}\left(\pi_\theta \| \pi_{\mathrm{ref}}\right)$ ensures $\pi_\theta$ does not deviate excessively from the reference policy $\pi_{\mathrm{ref}}$, with $\beta$ controlling the regularization strength. Unlike PPO, which relies on a value function for advantage estimation, GRPO leverages group statistics to compute $A_i$, eliminating the need for a separate critic model and reducing computational costs by 50\%. The algorithm iteratively updates $\pi_\theta$ by maximizing this objective, synchronizing parameters with $\pi_{\theta^{\text {old }}}$ for the next iteration, thus enabling efficient exploration of solution spaces to enhance reasoning capabilities, as demonstrated by notable improvements in benchmarks like MATH.
\begin{figure*}[t]
  \centering
  \label{fig:2}
  \includegraphics[width=\linewidth]{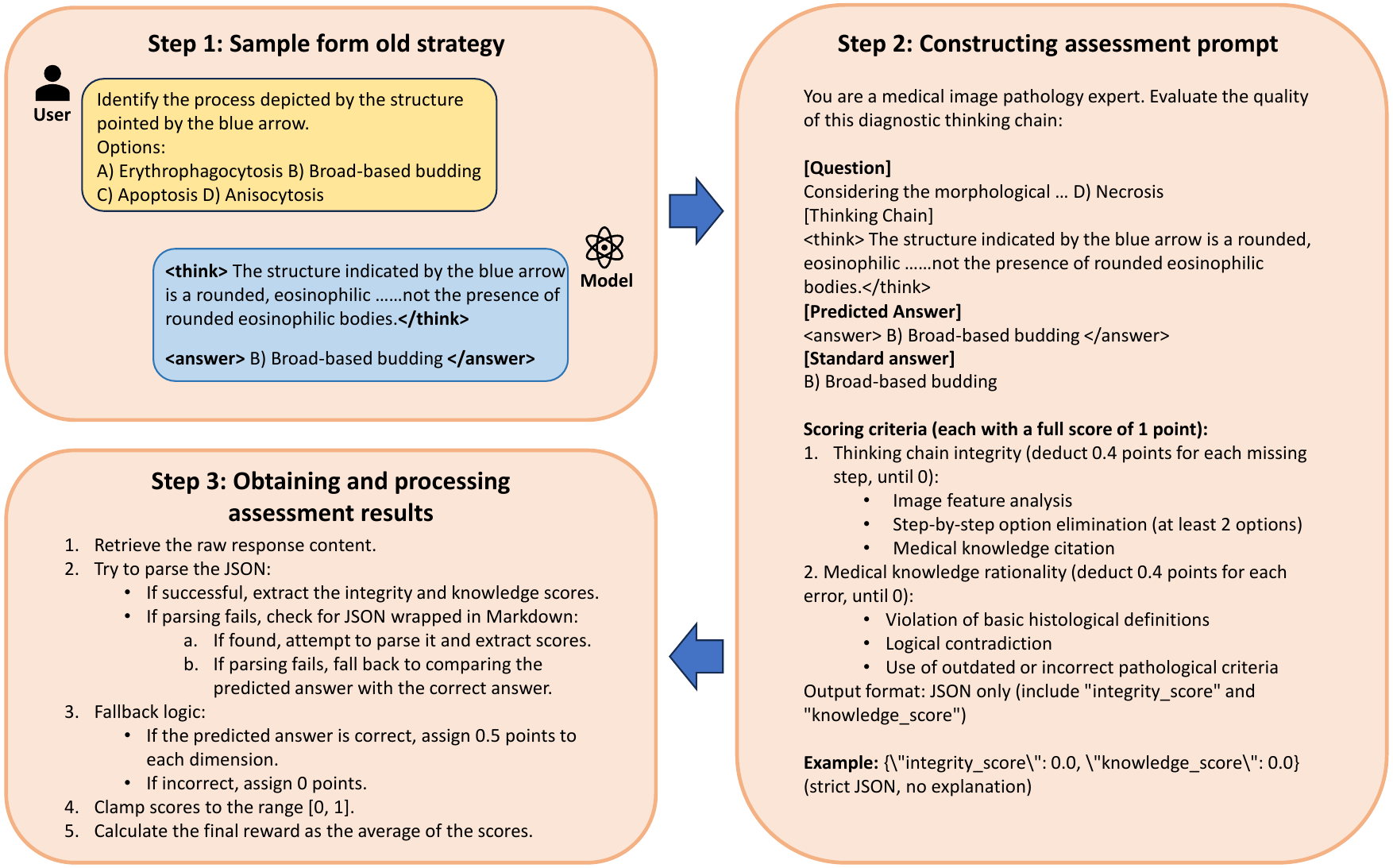}
  \caption{Pipeline for cross-modal procedural loss. First, sampling is performed on the old policy to obtain the model's generated thought process and final answers. These, along with the questions, generated content, and evaluation criteria, are then given to GPT-4o for review in terms of reasoning completeness and knowledge correctness, obtaining an Integrity score and Knowledge score. Lastly, GPT-4o's feedback is processed, including error handling and score normalization, using the average of the Integrity score and Knowledge score as the final reward.}
\end{figure*}

\subsection{Redesigning Reward Mechanisms for Interpretability}
During the training of Deepseek-R1\cite{deepseek-r1}, distinct reward mechanisms were tailored for different types of tasks. For tasks with well-defined standard answers or those that can be mechanically verified (such as mathematics, code, and logical reasoning), rule-based reward functions were utilized for optimization. For tasks lacking explicit rules or requiring subjective evaluation (such as creative writing and open-ended questions), a reward model capturing human preferences was introduced to address the limitations of rule-based rewards in unstructured tasks. This aims to balance reasoning capability with adaptability in general scenarios.

Similarly, to fulfill the requirements of process interpretability and result accuracy in reasoning within medical contexts, we designed a reward mechanism composed of cross-modal process rewards and result accuracy rewards. This mechanism ensures logical supervision of the reasoning process and constraints on result accuracy.

The result accuracy rewards follow the design of DeepseekMath\cite{GRPO}, comprising format rewards and accuracy rewards. The format reward requires the model to produce outputs in a specified format: explicitly articulating the reasoning process within a <think> tag and providing the final answer within an <answer> tag. Outputs that meet the format specifications receive a score of 1. In accuracy evaluation, the model must present a clear letter option and content within the <answer> tag, receiving a score of 1 only for completely correct answers.

However, test results based solely on format and accuracy rewards indicate that the quality of the generated reasoning chains is often suboptimal, frequently exhibiting logical leaps or misuse of medical arguments. Despite this, even low-quality reasoning chains can sometimes lead to correct results, which is detrimental to the model's ability to find effective strategies. We realized that for open-ended tasks such as medical reasoning, accuracy rewards alone are insufficient. External models are needed to supervise the reasoning process. Given the scarcity of high-quality reasoning process data in the medical field, we referenced Reinforcement Learning with AI Feedback (RLAIF)\cite{RLAIF1,constitution,MATRIX} and DeepSeek-R1\cite{deepseek-r1}'s training strategies to design a segmented approach.

We input images, questions, standard answers, model outputs, and prompts into the GPT-4o model, which evaluates them to generate cross-modal rewards. In the prompts, we specified the GPT-4o's evaluation criteria and response format. To prevent GPT-4o from returning biased formatting or irrelevant information, we designed a detailed process for handling its feedback.

Evaluation criteria include completeness of the reasoning chain and reasonableness of medical knowledge, each with a full score of 1 point. Completeness of the reasoning chain encompasses the analysis of image features, stepwise elimination of options, and reference to medical knowledge, with a deduction of 0.4 points for each missing step, with a minimum score of 0. Medical knowledge reasonableness involves breaches of basic histological definitions, logical contradictions, and the use of outdated or inappropriate pathological standards, with 0.4 points deducted for each infraction, also with a minimum score of 0. Figure 2\ref{fig:2} illustrates the detailed evaluation process.

\begin{figure}[t]
  \centering
  \includegraphics[width=\linewidth]{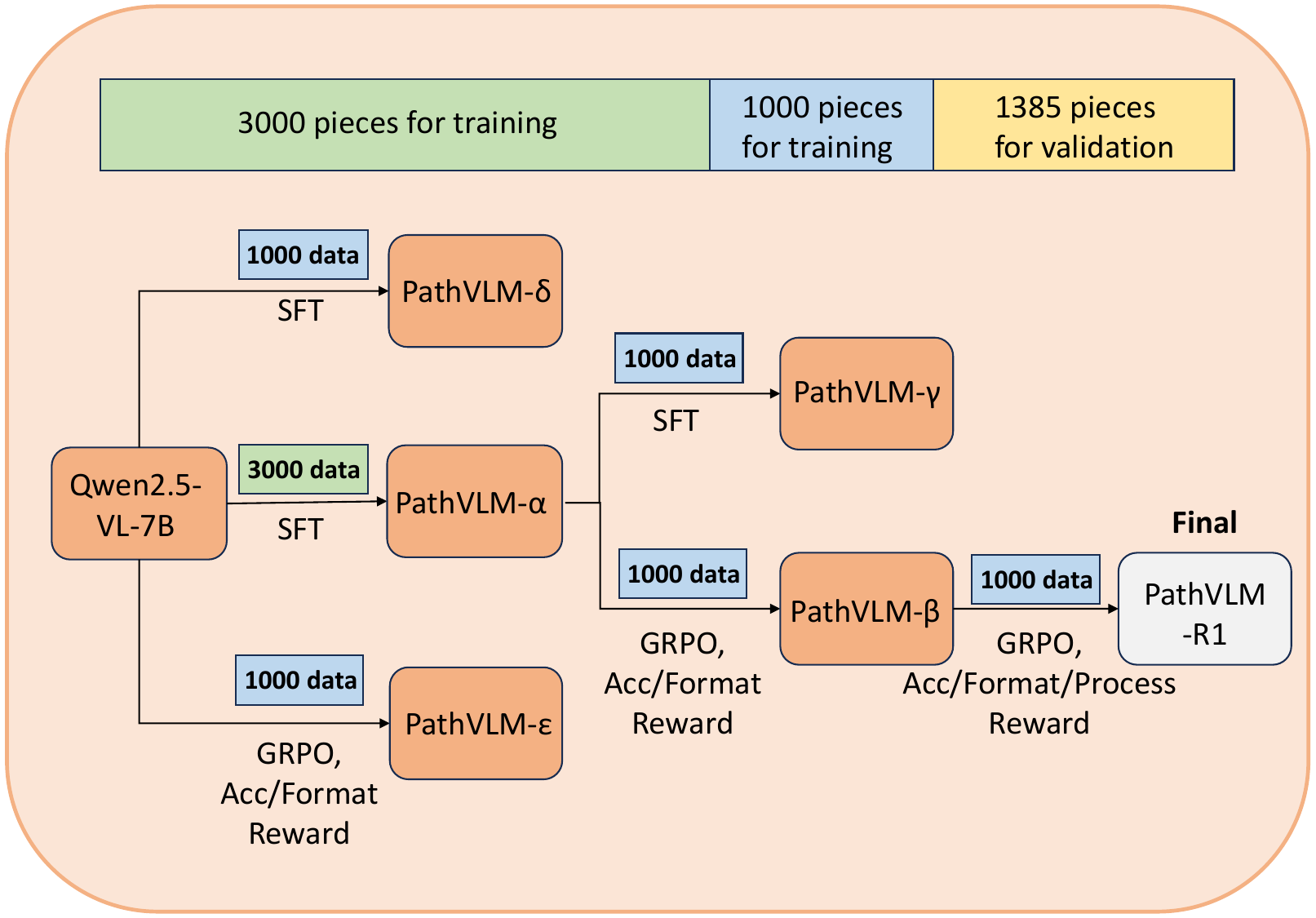}
  \caption{Various model variants generated during stage-wise training. The entire training set is divided into disjoint segments in the form of 3000, 1000, 1385, used for supervised fine-tuning, reinforcement learning, and its control group, as well as the final performance testing process, respectively. Using Qwen2.5-VL-7B as the base, the model Alpha is obtained with 3000 data points for supervised fine-tuning. Delta and Epsilon are obtained using 1000 data points for supervised fine-tuning/reinforcement learning. Gamma and Beta are obtained by continuing supervised fine-tuning/reinforcement learning with an additional 1000 data points on Alpha. Finally, adding cross-modal procedural losses to Beta results in the final model PathVLM-R1.}
  \label{fig3}
\end{figure}

\section{Experiment and Results}
\subsection{Experimental Setup}
\textbf{Dataset}
This study utilized the PathMMU\cite{pathmmu} dataset for experiments, which comprises 33,428 multimodal multiple-choice questions (Q\&A) and 24,067 pathological images. We selected the PubMed and EduContent subsets, totaling 5,385 entries for training and testing. Among these, 3,000 were used for training the Supervised Fine-Tuning base model, 1,000 for reinforcement learning training, and the remaining 1,385 for testing. Furthermore, to verify the model’s generalization capability, we selected ChestCT\cite{Chest_CT}, ISIC2020\cite{ISIC2020}, Retinal OCT-C8\cite{OCT-C8}, and Diabetic Retinopathy\cite{Diabetic_Retinopathy} from OmniMedVQA\cite{omnimedvqa} as out-of-domain data for testing.

\textbf{Implementation Details}
Experiments were conducted on a setup with four H20 GPUs (96GB of memory). The training base is Qwen2.5-VL-7B-Instruct\cite{qwen2.5}, with the visual encoder frozen. The parameter G is set to 4, and both the maximum prompt length and the maximum generation length are set to 1024.

The experimental procedure was divided into the following three phases:

The experimental procedure consists of the following three phases:

\begin{itemize}
\item Step 1: SFT Phase to Obtain the Alpha Model
Utilizing 3,000 data points to train the Qwen2.5-VL-7B-Instruct model for 18,000 steps until performance stabilizes, resulting in the new base model, Alpha.

\item Step 2: Reinforcement Learning Phase to Obtain the Beta Model
This is the first step in constructing reasoning capabilities. Using 1,000 newly introduced data points and applying rewards for accuracy and formatting only, Alpha is optimized for 2,000 steps until performance stabilizes, resulting in the Beta model.

\item Step 3: Final Optimization Using Cross-Modal Process Rewards
Continuing with the same 1,000 data points, combined with cross-modal process rewards and outcome accuracy rewards, the Beta model is further optimized for 2,000 steps until performance stabilizes, ultimately yielding the PathMMU model. We aim for the model to be faithful to the data distribution of the training set while avoiding potential reward hacking issues, thus choosing to further refine the reasoning process after initial model exploration.
\end{itemize}

\begin{table}[t]
\caption{In-domain comparison of different models and their accuracies (Acc. represents accuracy).}
\label{tab:1}
\centering

\begin{tabular}{lccc}
\hline
Method                               & Acc.          & Method                                  & Acc.             \\ \hline
\multirow{2}{*}{Qwen2.5-VL-3B}       & \multirow{2}{*}{44.48}       & \multirow{2}{*}{PathVLM-$\delta$-7B}     & \multirow{2}{*}{52.85}          \\
\multirow{2}{*}{Qwen2.5-VL-7B}       & \multirow{2}{*}{51.55}       & \multirow{2}{*}{PathVLM-$\epsilon$-7B}   & \multirow{2}{*}{56.10}           \\
\multirow{2}{*}{Qwen2.5-VL-32B}      & \multirow{2}{*}{54.20}        & \multirow{2}{*}{PathVLM-$\alpha$-3B}     & \multirow{2}{*}{54.58}          \\
\multirow{2}{*}{HuatuoGPT-Vision-7B} & \multirow{2}{*}{53.00}          & \multirow{2}{*}{PathVLM-$\beta$-3B}      & \multirow{2}{*}{56.89}          \\
\multirow{2}{*}{Gemini-1.5-Pro}      & \multirow{2}{*}{58.40}        & \multirow{2}{*}{PathVLM-$\gamma$-3B}     & \multirow{2}{*}{54.44}          \\
\multirow{2}{*}{PathVLM-$\alpha$-7B} & \multirow{2}{*}{57.59}       & \multirow{2}{*}{PathVLM-$\delta$-3B}     & \multirow{2}{*}{52.12}          \\
\multirow{2}{*}{PathVLM-$\beta$-7B}  & \multirow{2}{*}{{\ul 63.89}} & \multirow{2}{*}{PathVLM-$\epsilon$-3B}   & \multirow{2}{*}{49.60}           \\
\multirow{2}{*}{PathVLM-$\gamma$-7B} & \multirow{2}{*}{59.92}       & \multirow{2}{*}{\textbf{Ours(PathVLM-R1)}}  & \multirow{2}{*}{\textbf{65.55}} \\
\multirow{2}{*}{}  & \multirow{2}{*}{} \\
\hline
\end{tabular}

\vspace{0.2cm}  

\end{table}

Figure \ref{fig3} presents the model variants obtained during the training process.

To ensure the rigor of the experiment, we established the following control groups:

\begin{itemize}
\item Directly using 1,000 data points on the Qwen2.5-VL-7B-Instruct with accuracy and format rewards, optimized via GRPO, termed as model Epsilon, along with a supervised fine-tuning control group called model Delta.
\item Continued SFT training on the Alpha model, termed as model Gamma.
\end{itemize}

\textbf{Baselines and Evaluation Metrics}
Experiments were conducted on both in-domain pathological images and out-of-domain data. On the test set of in-domain data, PathVLM-R1's performance was compared against the following methods:
\begin{itemize}
\item Qwen2.5-VL series models, including 3B, 7B, and 32B versions;
\item HuatuoGPT-Vision-7B;
\item Gemini-1.5-Pro (using API);
\item Alpha, Beta, Gamma, Delta, Epsilon models, and their 3B versions.
\end{itemize}
In out-of-domain data testing, PathVLM-R1 was compared with:
\begin{itemize}
\item Qwen2.5-VL, including 3B and 7B versions;
\item Gamma model;
\item HuatuoGPT-Vision-7B.
\end{itemize}
The evaluation metric for model performance is accuracy, with a correct answer counted when the output's <answer> tag content perfectly matches the standard answer. To ensure test stability and reliability, diverse sampling is disabled by setting the do\_sample parameter to False.

\begin{table}[t]
\centering
\caption{Average number of tokens outputted by different models on In-domain data.}
\label{tab:2}
\begin{tabular}{lc}
\hline
Model         & Average number of tokens \\ \hline
Qwen2.5-VL-7B & 210.04                   \\
PathVLM-$\beta$-7B     & 80.31                    \\
PathVLM-$\epsilon$-7B     & 220.07                   \\
PathVLM-R1       & 94.99                    \\ \hline
\end{tabular}
\end{table}

\subsection{Experimental Results}

\textbf{Analysis of In-domain Data Results}
Table \ref{tab:1} summarizes the experimental results on in-domain data. When only supervised fine-tuning was applied, the accuracy of the 7B parameter model increased from 51.55\% to 57.59\% on the test set. Further improvements were achieved by applying reinforcement learning on the Alpha model with only accuracy and format rewards, boosting the model's accuracy to 63.89\%. By incorporating cross-modal process rewards, the accuracy further increased to 65.5\%, as shown in Figure \ref{fig:4}. This demonstrates the positive effect of improving the quality of the reasoning chain on final accuracy. This result also surpasses traditional visual language large models in the medical field such as Huatuo-GPT, as well as general visual language models like Qwen2.5-VL-32B and Gemini 1.5 Pro, which have significantly more parameters, as shown in Table \ref{tab:2}.
\begin{table}[t]
\centering
\caption{Out-of-domain comparison of different models and their accuracies}
\label{tab:3}
\begin{tabular}{lcccc}
\hline
\multirow{2}{*}{}   & \multicolumn{4}{c}{Out-of-Domain}                                 \\ \cline{2-5} 
        \centering Method                  & \parbox[c]{0.8cm}{\centering \vspace{0.1cm}Chest\\CT\vspace{0.1cm}}    & \parbox[c]{0.7cm}{\centering ISIC\\2020}    & \parbox[c]{1.1cm}{\centering \vspace{0.1cm}Retinal\\OCT-C8\vspace{0.1cm}} & \parbox[c]{1cm}{\centering \vspace{0.1cm}Diabetic\\Retinopathy\vspace{0.1cm}} \\ \hline
Qwen2.5-VL-3B             & 38.60        & 34.59       & \textbf{60.40}  & {\ul 64.20}           \\
Qwen2.5-VL-7B             & {\ul 48.60}  & {\ul 56.59} & {\ul 57.19}    & \textbf{70.80}        \\
HuatuoGPT-Vision-7B       & \textbf{51.00} & 34.20        & 48.19          & 62.00                   \\
PathVLM-$\gamma$          & 40.60        & 45.40        & 45.80           & 32.00                   \\
\textbf{Ours(PathVLM-R1)} & 44.00          & \textbf{80.00} & 49.20           & 59.80                 \\ \hline
\end{tabular}
\end{table}
In addition, we calculated the average output token count across different models on the test set. After 18,000 steps of supervised fine-tuning, there was a notable reduction in the output token count; however, this issue was alleviated by a further 2,000-step optimization with cross-modal process rewards.

Moreover, the performances of models Gamma, Delta, and Epsilon are noteworthy:
\begin{figure}[t]
  \centering
  \includegraphics[width=0.8\linewidth]{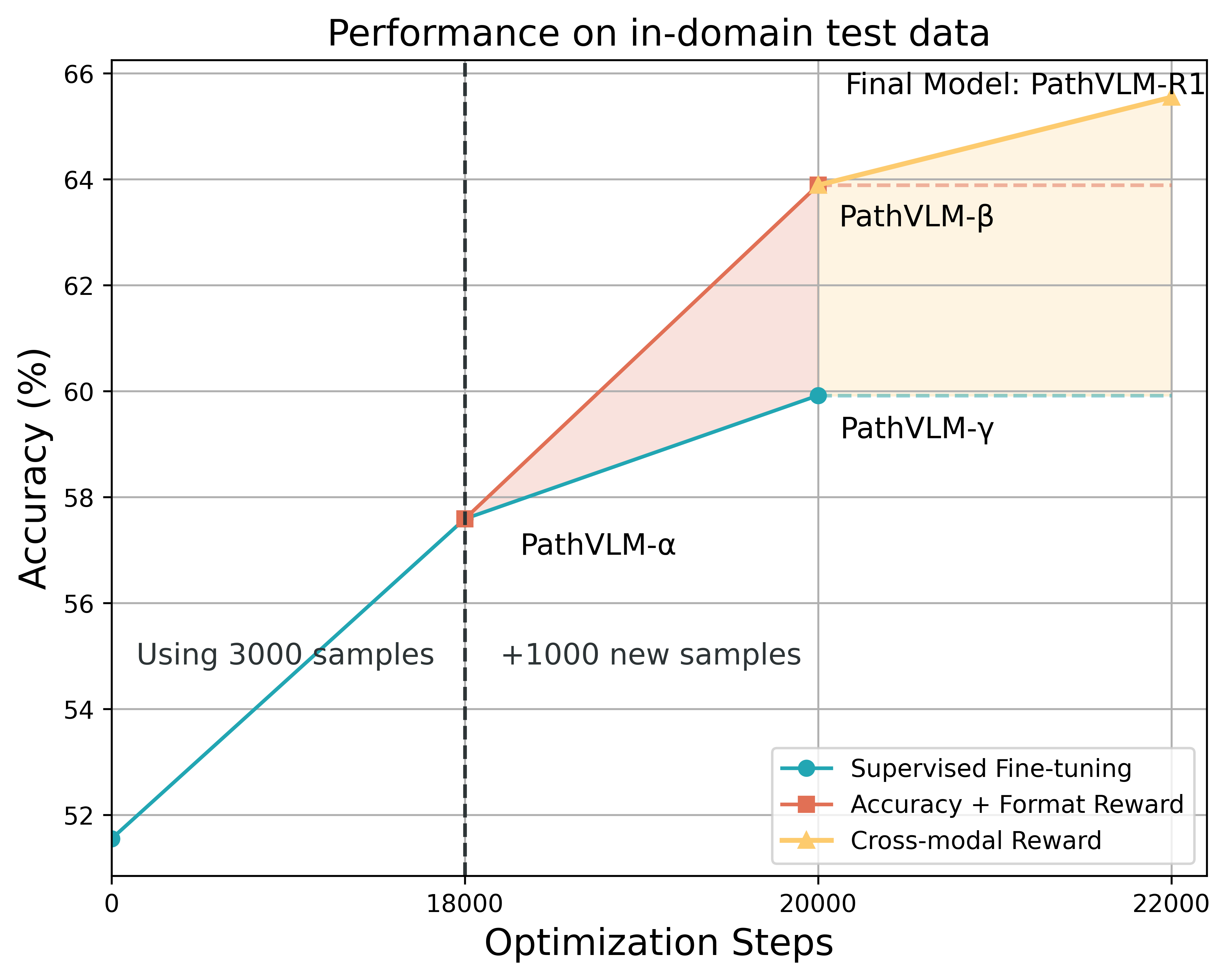}
  \caption{Changes in model accuracy across different optimization stages. The model's performance is significantly improved across all three training steps.}
  \label{fig:4}
\end{figure}
On the 7B parameter base, applying RL directly (model Epsilon) achieved an accuracy of only 56.10\%. After reaching this level, the model's performance began to fluctuate, likely due to a lack of domain knowledge, resulting in the learning of only superficial features or ineffective strategies.

Applying RL directly on the 7B parameter base (model Epsilon) outperformed the supervised fine-tuning baseline (model Delta), while on the 3B parameter base, the opposite was true.

After injecting domain knowledge via supervised fine-tuning followed by RL (Beta), the performance of the 3B parameter model surpassed the supervised fine-tuning baseline. By further applying our method on the 7B parameter base with only 1,000 new data points, the final model, PathVLM-R1, surpassed the supervised baseline (Gamma) by 7.96\%. In contrast, using the same 1,000 data points directly with RL resulted in only a 3.25\% increase.

These findings underscore the importance of model parameterization and domain knowledge in R1 model training: models with larger parameters have higher potential for performance and improvement, and pre-training with domain knowledge through supervised fine-tuning significantly enhances data efficiency and optimization in subsequent RL steps. This is particularly evident in domains like medical scenarios, which differ substantially from common modalities and lack manually annotated data. Other reasoning models for medical imaging modalities, such as Med-R1\cite{Med-R1} and MedVLM-R1\cite{medvlm-R1}, employ a rule-based reward function directly on the base model, optimized using the GRPO strategy. While this approach has also surpassed their foundational models and the supervised fine-tuning baseline, the results are suboptimal; MedVLM-R1 even failed in testing on pathology images. Our experimental results indicate that our post-training approach, which involves "pathological knowledge infusion and reasoning capability construction," can achieve superior performance and higher efficiency. Figure \ref{fig:1} illustrates the reasoning process and final results for the same question before and after adding cross-modal procedural supervision rewards. PathVLM-R1 achieves better outcomes due to its superior reasoning capabilities.

\textbf{Analysis of Out-of-domain Data Results}
Table \ref{tab:3} presents our test results on various out-of-domain modalities. Notably, PathVLM-R1 demonstrated a 23.41\% improvement over the baseline model on dermoscopic images. This modality's test data comprises macroscopic images representing visually apparent surface lesions, while our training data includes microscopic skin pathology slides. The results indicate effective transfer of domain knowledge acquired during training. Since HuatuoGPT-Vision-7B's training data encompass CT, MRI, and fundus images, which are absent from our training set, we do not hold an advantage over it in these out-domain modalities. Nevertheless, we still achieved performance superior to the supervised fine-tuning baseline, exhibiting impressive generalization capability and scalability potential. This highlights PathVLM-R1's adaptability and intrinsic value as a general model for the pathology domain.

\section{Conclusion}
This paper introduces PathVLM-R1, a visual-language reasoning model tailored for pathology images, capable of effectively enhancing reasoning capability and diagnostic accuracy. By employing a combined strategy of supervised fine-tuning and reinforcement learning based on Qwen2.5-VL-7B-Instruct, we achieved the infusion of pathological knowledge and structured construction of the reasoning process. Experimental results demonstrate that PathVLM-R1 significantly outperforms traditional models across various medical imaging domains, particularly excelling in pathology image question-answering tasks. Its dual-reward mechanism effectively integrates interpretability of the process with result accuracy, showcasing strong adaptability and generalization capabilities.

The study not only provides a new perspective for automated reasoning in pathology image diagnostics but also lays the groundwork for multi-scenario applications across other medical imaging modalities. Looking ahead, PathVLM-R1 is poised to further enhance performance in diverse application domains through richer data training and optimization strategies, advancing the automation of medical image analysis. We believe that as algorithms, computational power, and data evolve, PathVLM-R1 can be extended to broader clinical scenarios, contributing to the development of precision medicine and intelligent healthcare.

\bibliographystyle{unsrt}  
\bibliography{references}

\end{document}